\documentclass[a4paper,twoside]{article}

\usepackage{url}

\usepackage[table]{xcolor}

\usepackage{natbib}
\usepackage{hyperref}

\definecolor{darkblue}{rgb}{0, 0, 0.5}
\hypersetup{colorlinks=true, citecolor=black, linkcolor=darkblue, urlcolor=darkblue}

\renewcommand\cite{\citep}  % to get "(Author Year)" with natbib
\usepackage{epsfig}
\usepackage{subcaption}
\usepackage{calc}
\usepackage{amssymb}
\usepackage{amstext}
\usepackage{amsmath}
\usepackage{amsthm}
\usepackage{multicol}
\usepackage{pslatex}
\usepackage{apalike}
\usepackage{algorithm2e}
\usepackage[bottom]{footmisc}

\usepackage{SCITEPRESS}     % Please add other packages that you may need BEFORE the SCITEPRESS.sty package.

\begin{document}

\title{Diversifying Knowledge Enhancement of Biomedical Language Models using Adapter Modules and Knowledge Graphs}

\author{\authorname{Juraj Vladika, Alexander Fichtl, Florian Matthes 
%\orcidAuthor{0000-0000-0000-0000}
}
\affiliation{Department of Computer Science, Technical University of Munich, \\ Boltzmannstraße 3, 85748 Garching bei München, Germany}
\email{\{juraj.vladika, alexander.fichtl, matthes\}@tum.de}
}

%\author{\authorname{\textbf{Anonymous ICAART Submission}}}


\abstract{Recent advances in natural language processing (NLP) owe their success to pre-training language models on large amounts of unstructured data. Still, there is an increasing effort to combine the unstructured nature of LMs with structured knowledge and reasoning. Particularly in the rapidly evolving field of biomedical NLP, knowledge-enhanced language models (KELMs) have emerged as promising tools to bridge the gap between large language models and domain-specific knowledge, considering the available biomedical knowledge graphs (KGs) curated by experts over the decades. In this paper, we develop an approach that uses lightweight adapter modules to inject structured biomedical knowledge into pre-trained language models (PLMs). We use two large KGs, the biomedical knowledge system UMLS and the novel biochemical ontology OntoChem, with two prominent biomedical PLMs, PubMedBERT and BioLinkBERT. The approach includes partitioning knowledge graphs into smaller subgraphs, fine-tuning adapter modules for each subgraph, and combining the knowledge in a fusion layer. We test the performance on three downstream tasks: document classification, question answering, and natural language inference. We show that our methodology leads to performance improvements in several instances while keeping requirements in computing power low. Finally, we provide a detailed interpretation of the results and report valuable insights for future work.}

\onecolumn \maketitle \normalsize \setcounter{footnote}{0} \vfill

\section{\uppercase{Introduction}}
\label{sec:introduction}
The field of natural language processing (NLP) has been marked by impressive advancements in recent years. The appearance of new model architectures, including the emergence of generative transformers and pre-trained language models (PLMs), has brought along with it widespread usage and attention. Still, most of these models were trained on large amounts of web content, and while they excel at tasks in a general-purpose setting, there is still a performance gap when it comes to domain-specific challenges. 

One of these challenging domains is bio-medicine, which centers around the study of the human body, diseases, drugs, and treatments. Biomedical text is often characterized as highly complex because of its advanced terminology, which frequently includes names of chemical compounds, long-spanning relations, and other jargon not commonly used in everyday language. For NLP models trained on general corpora to work well in the biomedical domain, researchers have turned to transfer learning methods and domain adaption. The most common approach to domain adaptation is to continue the initial general pre-training of language models with data from domain-specific medical corpora. Examples of models adapted in this way are BioBERT \cite{biobert} and SciBERT \cite{scibert}, which drew the additional training data from biomedical and computer science research abstracts. Dropping the mixed-domain approach from previous frameworks, models like PubMedBERT \cite{blurb} and BioLinkBERT \cite{biolinkbert} were instead trained solely on PubMed research articles, with BioLinkBERT even leveraging links (citations) to other research articles.

While domain fine-tuning of whole PLMs has proven to increase the performance on downstream biomedical NLP tasks, additional pre-training can often be resource-intensive and infeasible for smaller research groups and situations where computing power is limited. A promising research direction has emerged in the form of knowledge-enhanced language models (KELMs) \cite{hu2023survey}. It refers to any set of methods that try to incorporate external knowledge into language models, usually by injecting it into the model's input, architecture, or output. In a sea of knowledge-enhancement methods, an especially interesting one is the utilization of adapters.

Broadly speaking, adapters are small bottleneck feed-forward layers inserted within each layer of a transformer-based language model  \cite{houlsby2019parameter,pfeiffer-etal-2020-adapterhub}. The small amount of additional parameters allows for the injection of new data or knowledge without requiring the whole model to be fine-tuned. Adapters plugged on top of large language models will often only have around 1\% of the number of training parameters compared to the transformer. The transformer model's learned parameters (weights) are frozen and left unchanged, and only the adapter is fine-tuned. Other than being lightweight on resources, this approach also helps avoid the problem of \textit{catastrophic forgetting}, where language models forget their existing knowledge from the pre-training corpora when they are fine-tuned on a new, smaller corpus \cite{ColonHernandez2021CombiningPL}.

This paper specifically focuses on using adapters to inject structured biomedical knowledge from large knowledge graphs into PLMs. We provide an overview of existing adapter approaches for the biomedical domain, as well as existing biomedical language models. We perform extensive experiments to test the performance of knowledge-enhanced, adapter-based biomedical language models on a number of representative biomedical classification tasks (document classification, question answering, natural language inference). We show that the model performance is improved in several instances on downstream tasks and provide a deeper look into the resulting change in model predictions. Finally, our experiments demonstrate that the OntoChem ontology \cite{irmer2013using}, which has not been used for knowledge enhancement yet, is a viable alternative to other prominent knowledge sources.

\section{\uppercase{Related work}}
\label{sec:related}

\subsection{Knowledge-Enhanced PLMs}

PLMs are trained on enormous corpora of training data, ranging from 3.3 billion tokens in the case of the original BERT \cite{devlin-etal-2019-bert}, all the way to 3.5 trillion tokens in the case of the recent Falcon-180B model \cite{falcon40b}. The power of the model architecture, combined with transfer learning, has led to these models showing impressive capabilities on most NLP tasks. While the textual data used for the model training is usually completely unstructured in nature, research has shown that models like BERT do encode, to some extent, syntactic structures, hierarchical concepts, and certain semantic conceptual relations \cite{rogers2021primer}. Still, other studies have shown weakness in modeling tasks dealing with structured knowledge, such as hyponymy relations \cite{ravichander-etal-2020-systematicity} or preserving the association between text and meaning \cite{diSciullo18}.

In most cases, the knowledge we find and gather, especially scientific knowledge, can be represented in a structured manner. This is the underlying idea of knowledge graphs (KGs), a data structure that models concepts (entities) and relations between them in a graph-like format \cite{ji2021survey}. KGs have been used in the field of NLP to enhance the performance of NLP models in many downstream NLP tasks \cite{schneider-etal-2022-decade}. There are multiple ways to combine KGs with PLMs. The knowledge triples from KGs can be embedded as vector representations such as TransE \cite{wang2014knowledge} or TuckER \cite{balazevic-etal-2019-tucker} and then combined with the vectors encoding text. Alternatively, the triples from KGs can be converted to sentences, and, in turn, these textual representations can then be used to fine-tune PLMs in the same way as with any other text. This approach was followed by COMET \cite{bosselut-etal-2019-comet}, which utilized the knowledge graph ConceptNet \cite{speer2017conceptnet} to enhance the performance on commonsense reasoning tasks. Besides knowledge graphs, lexicons are sometimes used for knowledge enhancement \cite{Hoang22}.

While there are numerous ways to inject structured knowledge into PLMs such as adding it to the input and output of models \cite{Wei21}, an especially promising approach is adding adapters to the architecture of the model \cite{ColonHernandez2021CombiningPL}. Adapters are small layers that are inserted within a language model and are subsequently fine-tuned to a specific task. The major benefit of adapters is that they add a minimal amount of additional parameters, thus significantly reducing the needed training time. Combined with freezing original model weights, adapters can avoid catastrophic forgetting, where the PLM's performance deteriorates when all of its weights are fine-tuned with a new knowledge source. Adapters have been used for numerous purposes such as learning hierarchical representation \cite{chronopoulou-etal-2022-efficient}, transferring models trained on English to low-resource languages \cite{Wang2021EfficientTT}, and in the domain of efficient transformers as low-rank adapters (LoRA) \cite{hu2022lora}. General knowledge-enhanced PLMs utilizing adapters include, for example, KnowBERT \cite{knowbert} and K-Adapter \cite{kadapter}. A practical tool emerged that combines well-known adapter architectures in one place, called AdapterHub \cite{pfeiffer-etal-2020-adapterhub}.

\subsection{Biomedical Knowledge-Enhanced PLMs}

A major focus of knowledge enhancement in PLMs is in domain adaption to expert domains such as the biomedical domain. So far, most of the advancements have focused on utilizing the knowledge graph UMLS \cite{bodenreider2004unified} for this purpose.  Examples include BERT-MK \cite{he-etal-2020-bert} and KeBioLM \cite{yuan-etal-2021-improving}, which both fine-tune the whole weights of the base language model by using masked language modeling of triples from UMLS. Biomedical PLMs can then be used for various NLP tasks, such as biomedical text summarization \cite{abacha2021overview}, named entity recognition \cite{sung2022bern2}, medical fact-checking \cite{vladika-matthes-2023-scientific}, information retrieval \cite{luo2022improving}, or health question answering \cite{vladika2023healthfc}.

There are also existing approaches using adapters for biomedical knowledge enhancement. Representative works are DAKI \cite{daki}, which fine-tunes the adapters with entity prediction task, and KEBLM \cite{keblm}, which fine-tunes the adapters on three different knowledge types from UMLS and PubChem \cite{kim2019pubchem}, namely entity descriptions, entity-entity relations, and entity synonyms. The most similar approach to ours and a direct inspiration was the Mixture-of-Partitions (MoP) approach \cite{meng-etal-2021-mixture}, where the adapters were fine-tuned on smaller subgraphs of UMLS.

\begin{figure}[htpb]
  \centering
  \includegraphics[width=0.99\linewidth]{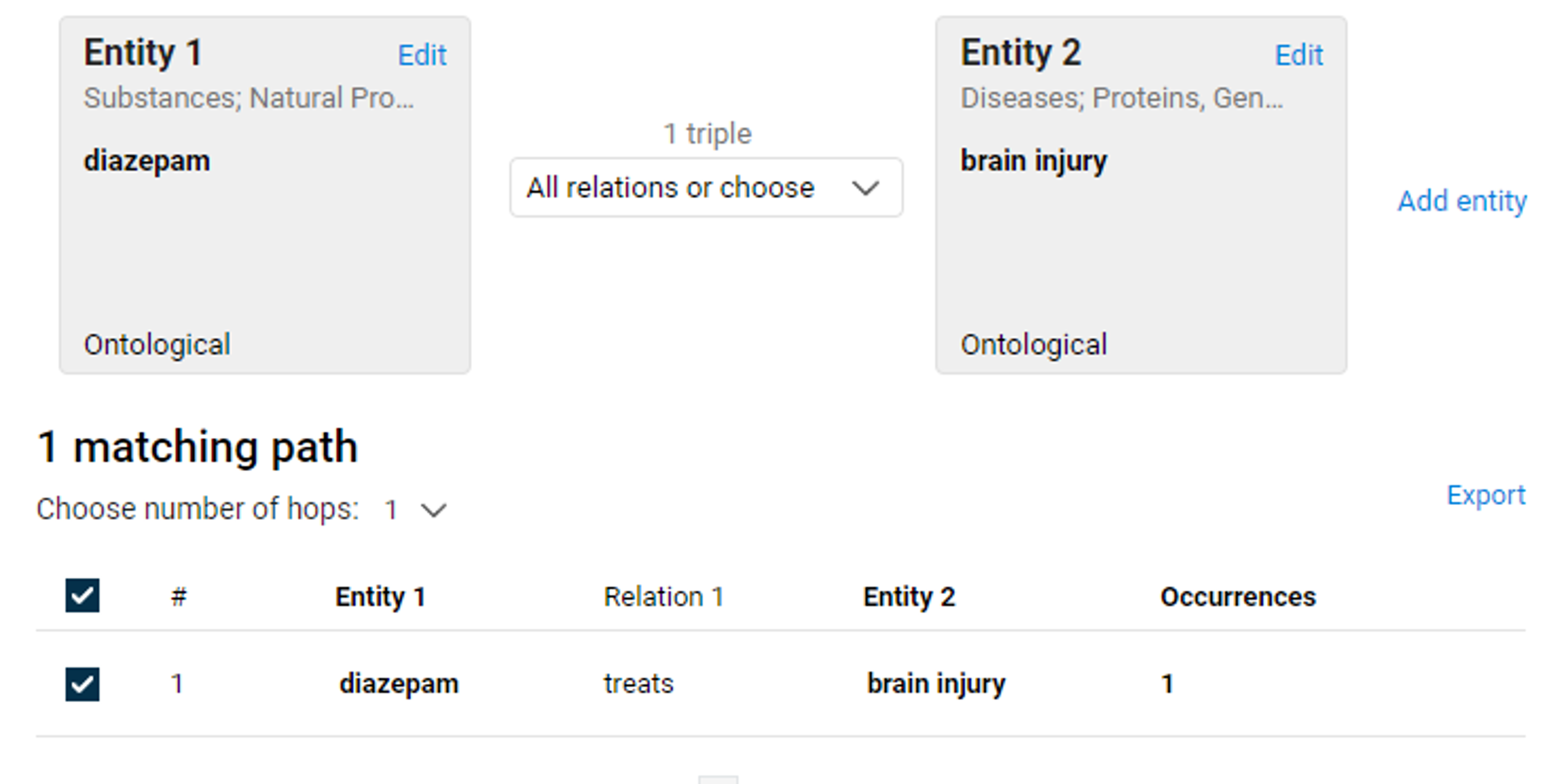}
  \caption[Caption]{Triplet from the OntoChem Fact Finder\footnotemark}
  \label{fig:ontochem}
\end{figure}

In essence,\footnotetext{\url{https://sciwalker.com/analytics/factfinder}} our work builds on the present foundations of adapter-based biomedical models and uses the yet unexplored knowledge graph OntoChem, which is rich with chemical knowledge. For our experiments, we use the well-known biomedical PLM PubMedBERT as well as the yet unexplored but powerful BioLinkBERT base model. Following the suggestions of \cite{meng-etal-2021-mixture}, we use only the triplets corresponding to the 20 most frequent relations of OntoChem for the knowledge injection. An example of an OntoChem triplet can be seen in Figure \ref{fig:ontochem}. Finally, we provide a deeper qualitative analysis of learned structured knowledge on a specific dataset. Notably, our work achieves the SOTA (averaged) performance on the question-answering BioASQ-7b dataset.

\section{\uppercase{Methodology}}
\begin{figure*}[htpb]
  \centering
  \includegraphics[width=0.99\linewidth]{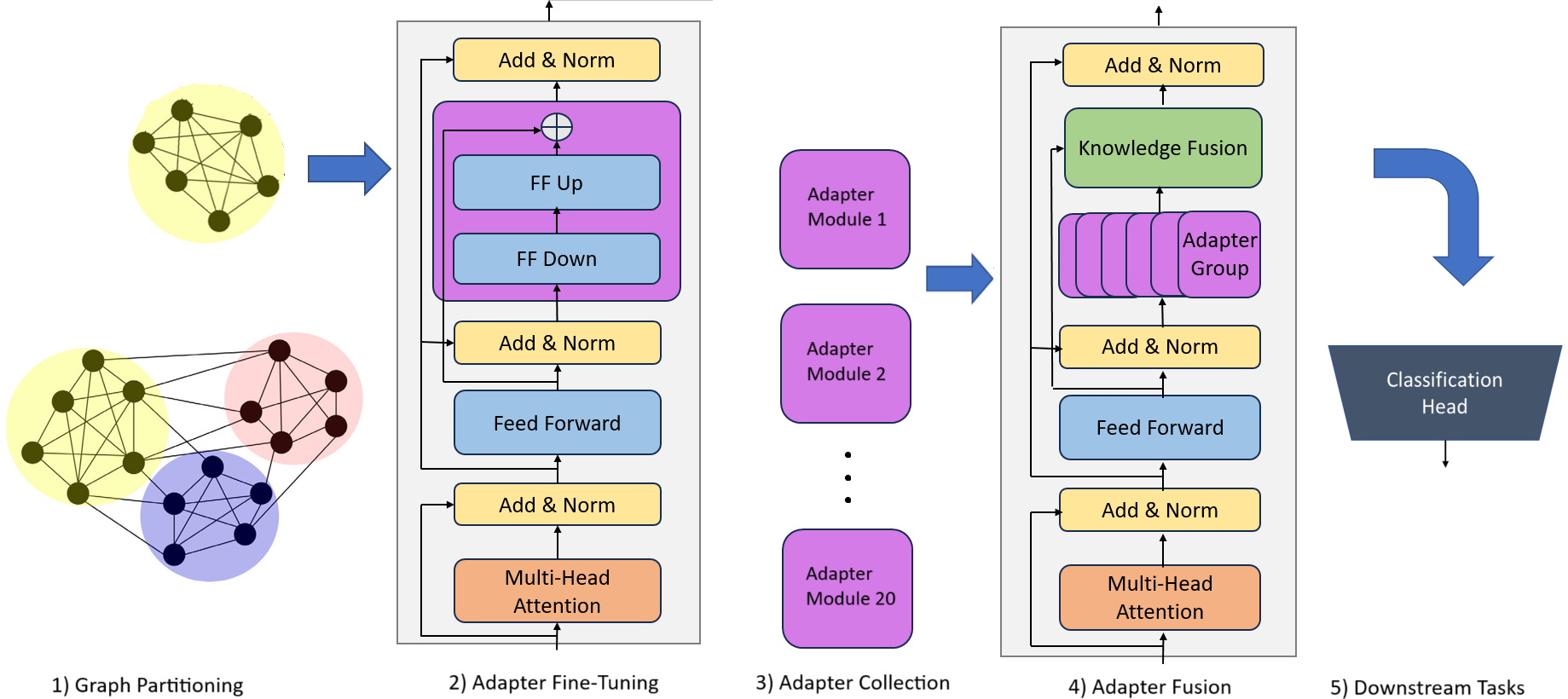}
  \caption{Methodology used to construct the final model and run the experiments}
  \label{fig:methodology}
\end{figure*}

In this section, we will explain the training methodology we used for the experiments in this paper. It is depicted in Figure \ref{fig:methodology}.

\subsection{Knowledge Graph Representation}
A central element of our method is the knowledge graph (KG). This KG is a structured representation of information denoted as a collection of ordered triples \cite{ji2021survey}. We denote these triples as $(s, r, o)$, where \textit{s} is a subject, \textit{r} is a relation, and \textit{o} is an object. Both \textit{s} and \textit{o} are entities that come from an entity set \textit{E}, while relations come from a relation set \textit{R}. Each entity and relation in the KG is associated with its corresponding textual surface form. This form can take the shape of a single word or a compound term (e.g., for names of chemicals) or even a concise phrase, especially in case of relations. This textual association is critical as it bridges the gap between the structured KG and natural language, allowing for easier injection of KG knowledge into the language models and associated fine-tuning.

The primary objective is to enhance the capabilities of a pre-trained language model, denoted as \textit{LM}, by integrating the knowledge contained within the KG. To achieve this, we need a training objective that effectively incorporates the KG knowledge into the model. Most encoder-only PLMs based on the original BERT use masked language modeling (MLM) as one of its pre-training objectives. This task consists of masking a certain word in a given sentence and having the model predict which word would fit the best in the place of the masked token. We follow the established approach of using an entity prediction objective, where we mask one of the entities and have the model predict which token would best fit. In this way, the model incorporates the structured knowledge of $(s, r, o)$ triples into its internal weights. 

%Several approaches can be used to formulate this training objective. These include relation classification, entity linking, next sentence prediction, and entity prediction. In this particular paper, the focus is on entity prediction, which is a widely used and recognized approach for enhancing language models with external knowledge.

\subsection{Graph Partitioning}
During the prediction of the masked token, the model produces a probability distribution (with a softmax function) over all of the entities from KG's entity set \textit{E}. Considering the massive size of the biomedical KGs we use in the paper, computing the softmax function over all its entities carries a lot of computation complexity. This issue can significantly slow down model training and inference. To bridge this challenge, some approaches have been suggested in the literature. We opt for the approach of \cite{meng-etal-2021-mixture}, which involves partitioning the KG into smaller subgraphs, which are then trained on independently, and later, their knowledge combined to unified knowledge representations. 

The process of dividing a KG yields smaller subgraphs that we denote as $G_1, G_2, ..., G_k$. We set $k$ in final experiments to be $20$, following empirical observations and previous literature, which balances efficiency and graph coverage well. Ideally, these $20$ smaller subgraphs should be almost equal-sized components, meaning nodes are balanced across partitions. Additionally, the capacity of edges between different components should be minimized to maximize the retention of factual knowledge. This is a problem known as \textit{balanced graph partitioning} and is known to be an NP-complete problem \cite{andreev2004}. Several good approximations have been developed to determine the exact solution despite its computational complexity. We opt for the METIS algorithm \cite{karypis1997metis}, which was used in other works dealing with large-scale KG partitioning \cite{zheng2021}. 
%The algorithm works as follows...

%The key to making MoP effective lies in the process of graph partitioning, which essentially means dividing the nodes (entities) of the KG into mutually exclusive groups (sub-graphs). This step is crucial for two reasons. Firstly, it facilitates data parallelism, allowing for efficient processing of the KG across multiple computational units. Secondly, it helps in controlling the overall computational burden, which is essential when dealing with large KGs.

\subsection{Adapter Module Learning}
Once the KG is appropriately partitioned, the process of fine-tuning the LM can be started. We deploy adapter modules for this purpose. As mentioned previously, adapters are newly initialized feed-forward networks inserted between the transformer model's layers. Notably, the training of adapter modules does not require fine-tuning the existing parameters of the pre-trained model. Instead, it focuses solely on updating the parameters within the adapters. This strategy ensures that the pre-trained model's core knowledge remains intact while enabling the model to specialize in the biomedical domain by adapting to the specific knowledge contained in the KG.

There are multiple adapter module configurations, such as \cite{houlsby2019parameter} and \cite{bapna-firat-2019-simple}. The adapter module configuration used in the paper is based on the one by \cite{adapterfusion}, the so-called Pfeiffer architecture. In this configuration, only one adapter module is added as a down-projection and up-projection, unlike the Houlsby architecture, where there are two projections. While the Houlsby architecture has more learning capacity, it comes with training and inference speed costs. Previous studies showed no significant difference in performance between the model architectures, making Pfeiffer architecture a very lightweight choice that brings powerful learning capabilities. 

As already mentioned, masked language modeling is used to fine-tune the adapter modules. More precisely, it is a task of entity prediction since a missing entity from the graph triple is being predicted. Given a subgraph $G_k$ and its triples $(s, r, o)$, each triple has a textual representation. The object entity \textit{o} is removed from each triple, and the remaining two elements of the triple are transformed into a textual representation like: "[CLS] s [SEP] r [SEP]". The adapter module is then trained to predict the missing object entity using the representation of the [CLS] token. The parameters of the adapter module are optimized by minimizing the cross-entropy loss.

%During the fine- tuning of downstream tasks, both the parameters of ADAPTER and pre-trained LM will be updated  
%The relations provided by OntoChem are unique to the type of entities that the relation connects, so there can be several types of "induces" relations. For example, there is one possible relation for a "substance" as a subject and a "disease" as an object ([substance] induces [disease]), and another one for a "physiology" as a subject and a "disease" as an object ([physiology] induces [disease]).  

\subsection{Knowledge Fusion}
Finally, with a set of knowledge-encapsulated adapter modules at hand, we need to fuse their knowledge together into a final representation. For this, we use the so-called AdapterFusion mixture layers \cite{adapterfusion}. These layers serve the purpose of combining knowledge from various adapters to enhance the model's performance on downstream tasks. It is a relatively recent approach designed to effectively learn how to combine information from a set of task-specific adapters. It does so by employing a softmax attention mechanism that assigns contextual mixture weights over the adapters. These weights are then used to predict task labels in the final layer. The composition of these layers and their interactions ultimately contribute to the model's ability to generalize and perform well on a range of tasks.

%AdapterFusion can be seen as a part of a broader family of models that incorporate mixture-of-experts concepts. It's closely related to the sparsely-gated Mixture-of-Experts layer. Additionally, it offers flexibility in terms of using mechanisms like Gumbel-Softmax for obtaining more discrete or continuous mixture weights. However, in their experiments, the authors found that AdapterFusion outperforms these alternatives, demonstrating its effectiveness in fusing knowledge from multiple sources.

%In summary, this paper presents a comprehensive approach for enhancing language models in the biomedical domain by leveraging the rich knowledge contained within a KG. The method involves graph partitioning, ADAPTER modules, and AdapterFusion mixture layers, all of which contribute to the model's ability to understand and generate text specific to biomedical topics. This work addresses the critical challenge of incorporating structured knowledge into language models and provides valuable insights for advancing research in the field.

%Explain MedNLI, MedQA, BioASQ, PubMedQA.

\section{\uppercase{Experiments}}
\label{sec:experiments}

In this section, we describe our approach to leveraging data from OntoChem's SciWalker platform together with adapters to improve existing approaches to biomedical KELMs. For reproducibility, we made the code for the experiment runs available on GitHub.\footnote{\url{https://github.com/alexander-fichtl/diversifying_KELMs.git}}

%Done: Change MedNLI BLURB
\subsection{Datasets}
All of our datasets, with the exception of MedNLI, originated from a collection of common biomedical NLP tasks known as BLURB -- Biomedical Language Understanding and Reasoning Benchmark.\footnote{\url{https://microsoft.github.io/BLURB/index.html}} Inspired by a similar suite of tasks for general-purpose natural language understanding (NLU) known as GLUE \cite{wang-etal-2018-glue}, BLURB covers a wide-range of tasks related to biomedical NLU. This means no tasks include text generation and are all essentially classification tasks, which makes them convenient to evaluate with common classification metrics such as precision, recall, accuracy, and F1 score. The four datasets are described in continuation. 

\noindent \textbf{MedNLI} \cite{romanov-shivade-2018-lessons} is a dataset for natural language inference (NLI). It consists of 14,049 unique sentence pairs, where one sentence is a hypothesis, and the other one is a premise. The task is to infer whether the premise entails the hypothesis, contradicts it, or is in a neutral relation with respect to it. The premises were collected from MIMIC-III \cite{johnson2016mimic}, the largest repository of publicly available clinical data (patient notes).

\noindent \textbf{BioASQ-7b} \cite{10.1007/978-3-030-43887-6_51} is a biomedical question answering (QA) benchmark dataset containing questions in English, along with golden standard (reference) answers and related material. It has been designed to reflect real information needs of biomedical experts. Other than only exact answers, the BioASQ dataset also includes ideal answers (summaries). Researchers working on paraphrasing and textual entailment can also measure the degree to which their methods improve the performance of biomedical QA systems. The dataset is a part of the ongoing shared challenge with the same name \cite{tsatsaronis2015overview}, while our dataset (7b) is from the 2019 challenge. 

\noindent \textbf{PubMedQA} \cite{jin-etal-2019-pubmedqa} is a different QA dataset collected from PubMed abstracts, the largest collection of biomedical research papers \cite{white2020pubmed}. The task of PubMedQA is to answer research questions with yes/no/maybe using the corresponding abstracts. The dataset has 1,000 expert-annotated instances of question-answer pairs. Each PubMedQA instance is composed of a question, a context (abstract without the conclusion), a long answer (conclusion of the abstract), and a yes/no/maybe label that summarizes the conclusion.

\noindent The \textbf{Hallmarks of Cancer} (HOC) Corpus \cite{baker2015automatic} consists of 1852 PubMed publication abstracts manually annotated by experts according to a taxonomy. The taxonomy consists of 37 classes in a hierarchy. Zero or more class labels are assigned to each sentence in the corpus. These hallmarks refer to the alterations in cell behavior that characterize the cancer cell. Proposed as a strategy to capture the complexity of cancer in a few basic principles, it provides an organized framework comprising of ten hallmarks \cite{10.1093/bioinformatics/btx454}.

\begin{table*}[htpb]
    \centering
    \scriptsize
    \begin{tabular}{p{3.1cm}l|p{3.1cm}l|p{4.2cm}l}
    \hline
        \textbf{UMLS20} & \textbf{\#Triples} & \textbf{Onto20Fused} & \textbf{\#Triples} & \textbf{Onto20Type} & \textbf{\#Triples} \\ \hline
        has finding site & 367,237 & relates to & 708,076 & [protein] relates to [disease] & 295,841 \\ 
        has method & 275,398 & induces & 502,512 & [substance] induces [physiology] & 282,721 \\ 
        has associated morphology & 269729 & modulates & 326,534 & [food] contains [compound] & 269,211  \\ 
        has procedure site & 129,686 & treats & 225,279 & [substance] treats [disease] & 247,348  \\ 
        possibly equivalent to & 91,446 & inhibits & 219,720 & [biomarker] of [disease] & 205,604  \\ 
        has causative agent & 86,780 & is analyzed by & 195,291 & [substance] is analyzed by [method] & 130,275  \\ 
        interprets & 84,533 & produces & 173,979 & [plant] produces [compound] & 102,270  \\ 
        has direct procedure site & 83,749 & increases activity of & 148,673 & [protein] induces [physiology] & 85,411 \\ 
        has active ingredient & 63,792 & contains & 133,241 & [compound] increases activity of [protein] & 85,196 \\ 
        has pathological process & 54,639 & increases & 110,803 & [compound] decreases activity of [protein] & 72,311  \\ 
        has occurrence & 40,154 & detects & 93,373 & [substance] inhibits [physiology] & 68,728  \\ 
        has dose form & 30,940 & decreases activity of & 85,425 & [protein] is a [biomarker] & 65,558  \\ 
        has direct morphology & 29,667 & prevents & 82,574 & [anatomy] produces [protein] & 64,206  \\ 
        has intent & 25,907 & increases expression of & 80,771 & [substance] prevents [disease] & 60,260  \\ 
        has interpretation & 24,624 & expresses & 62,142 & [protein] induces [disease] & 59,577  \\ 
        has direct substance & 23,042 & attenuates & 54,865 & [substance] modulates [protein] & 54,533  \\ 
        has direct device & 17,726 & decreases expression of & 51,152 & [protein] is analyzed by [method] & 54,250  \\ 
        moved to & 17,507 & binds to & 49,206 & [method] treats [disease] & 35,768  \\ 
        has temporal context & 17,195 & is a & 47,435 & [method] detects [physiology] & 33,504 \\
        has subject relationship context & 16,926 & affects expression of & 37,399 & [protein] modulates [physiology] & 24,332 \\ \hline
        \textbf{Total} & \textbf{1,750,677} & & \textbf{3,388,450} &  & \textbf{2,296,904} \\ \hline

    \end{tabular}
    \caption{Twenty most common relations in each of the three KGs used in the experiments}
    \label{tab:top20relations}
\end{table*}

\subsection{Knowledge Sources}
The Unified Medical Language System (\textbf{UMLS}) is a set of resources and tools developed by the US National Library of Medicine (NLM) to facilitate the integration and retrieval of biomedical and clinical information from various sources \cite{bodenreider2004unified}. Created in 1986 and continuously developed over the decades, it can be viewed as a comprehensive thesaurus and ontology of biomedical concepts, making it easier to connect and use medical terminology in research, clinical practice, and healthcare information systems. We use the most recent \textit{SNOMED CT, US Edition} vocabulary from September 2023.\footnote{\url{https://www.nlm.nih.gov/healthit/snomedct/us_edition.html}}

The second knowledge graph, more precisely ontology, that we use, is the \textbf{OntoChem} Ontology \cite{irmer2013using}.  The ontology contains more than 900 complex relationships between two or more named entities. Entities include chemical compounds, diseases, drug combinations, chemical reactions, biological activities, adverse reactions, etc. Relationships can be downloaded as RDF files. The data originates from MedLine,\footnote{\url{https://www.nlm.nih.gov/medline/index.html}} a bibliographic database from the US National Library of Medicine’s (NLM), that contains more than 30 million journal articles focusing on medicine and life sciences. The KG triples can be interactively queried and also downloaded from the SciWalker platform with the Fact Finder tool.\footnote{\url{https://sciwalker.com/analytics/factfinder}}

\subsection{KG Subsets}
The versions of the KGs from the two knowledge sources we use in this work are subsets of their respective full KGs -- we use versions including only the top 20 most common relations. This was done to increase the efficiency of training but also because initial experiments showed this smaller version does not hurt the performance on downstream tasks. For UMLS, the list of most common relations was taken from MoP and SNOMED, a systematically organized collection of medical terms providing codes, terms, synonyms and definitions used in clinical documentation and reporting. We label this KG as \textbf{UMLS20}.

The relations provided by OntoChem are unique to the type of entities that the relation connects, so there can be several types of the same relation. For example, the relation "\textit{induces}" can have a "\textit{substance}" as a subject and a "disease" as an object, so the full relation becomes "\textit{[substance] induces [disease}]", while another one is with a "\textit{physiology}" as a subject and a "\textit{disease}" as an object, producing "\textit{[physiology] induces [disease]}". To test the performance between these two types, we produce both a KG with top 20 fused relations (independent of entity types) and with top 20 typed relations (dependent on entity types). We call these two KGs \textbf{Onto20Fused} and \textbf{Onto20Type}. 

The top 20 relations in each of the three KGs is shown in Table \ref{tab:top20relations}. This also gives a good insight into what kind of structured knowledge is actually contained in these manually curated biomedical knowledge bases. While there are certain overlaps between top relations UMLS and OntoChem, a lot of them refer to different types of interactions between entities. Therefore, a promising research avenue that we did not explore in this work would be to merge these two knowledge bases into a unified KG and use both to fine-tune the adapters.

\subsection{Setup}
Task-specific fine-tuning is carried out for the four chosen benchmark downstream tasks. We aligned our hyperparameters with the settings recommended by the BLURB creators \cite{blurb}: We deploy the Adam optimizer \cite{zhang2018improved} alongside the typical slanted triangular learning rate schedule, with a warm-up for the initial 10 percent of steps and a cool-down for the subsequent 90 percent, and set the dropout probability at 0.1. Furthermore, we followed \cite{adapterfusion} and  \cite{meng-etal-2021-mixture} by introducing mixture layers and AdapterFusion to route valuable knowledge from the adapters to downstream tasks automatically. Given the random initialization of the task-specific model and dropout, outcomes can fluctuate based on different random seeds, particularly for the small PubMedQA and BioASQ7b datasets. For a more accurate representation, we present average results from ten iterations for BioASQ7b and PubMedQA, five iterations HoC, and three for MedNLI, as done in related biomedical NLP papers benchmarking these tasks. 

The training was carried out on Google Colab, with V100 and T4 GPUs provided on the platform. Specific hyperparameters and settings used in our experiments are shown in Table \ref{tab:hyperparameters}. Run seeds are reported on GitHub.

\begin{table}[htpb]
    \centering
    \scriptsize
    \begin{tabular}{c|cccc}
        \textbf{Setting/Task} & \textbf{HoC} & \textbf{PubMedQA} & \textbf{BioASQ7b} & \textbf{MedNLI} \\
        \hline
        repeat runs & 5 & 10 & 10 & 3 \\
        epochs & 20 & 30 & 25 & 20 \\
        patience & 3 & 4 & 5 & 3 \\
        batch size & 16 & 4 & 4 & 8 \\
        learning rate & 1e-5 & 0.5e-5 & 0.5e-5 & 0.5e-5 \\
        max.\ seq.\ len.\ & 128 & 512 & 512 & 256 \\
    \end{tabular}
    \caption{Settings and hyperparameters used for training each of the datasets of the downstream tasks}
    \label{tab:hyperparameters}
\end{table}

\section{\uppercase{Results}}

This section describes the detailed experiment results. We provide both a numerical analysis and a qualitative analysis of the results.

\begin{table*}[htpb]
\centering
\begin{tabular}{lllll}
\hline
\textbf{$\downarrow$ model|dataset $\rightarrow$} & \textbf{HoC} & \textbf{PubMedQA} & \textbf{BioASQ7b} & \textbf{MedNLI}\\ \hline
\textbf{\textit{SciBERT-base}}                           & \textit{80.52}$_{\pm0.60}$                             & \textit{57.38}$_{\pm4.22}$                             & \textit{75.93}$_{\pm4.20}$                     & \textit{81.19}$_{\pm0.54}$     \\
\textbf{\qquad+ \textit{MoP}}                              & \textit{81.79}$^\dagger_{\pm0.66}\uparrow$           & \textit{54.66}$_{\pm3.10}$                             & \textit{78.50}$^\dagger_{\pm4.06}\uparrow$     & \textit{81.20}$_{\pm0.37}$     \\
\textbf{\qquad+ \textit{KEBLM}}                            & /                                           & \textit{59.0}                                          & /                                     & \textit{82.14}     \\ \hdashline
\textbf{\textit{BioBERT-base}}                           & \textit{81.41}$_{\pm0.59}$                             & \textit{60.24}$_{\pm2.32}$                             & \textit{77.50}$_{\pm2.92}$                     & \textit{82.42}$_{\pm0.59}$    \\
\textbf{\qquad+ \textit{MoP} }                             & \textit{82.53}$^\dagger_{\pm1.08}\uparrow$           & \textit{61.04}$_{\pm4.81}\uparrow$                     & \textit{80.79}$^\dagger_{\pm4.40}\uparrow$     & \textit{82.93}$_{\pm0.55}\uparrow$    \\ 
\textbf{\qquad+ \textit{KEBLM}}                            & /                                           & \textbf{\textit{68.00}} $\uparrow$                                & /                                     & \textit{84.24} $\uparrow$  \\
\textbf{\qquad+ \textit{DAKI}}                             & /                                           & /                                             & /                                     & \textit{83.41} $\uparrow$     \\ \hline
\textbf{PubMedBERT-base}                        & \textit{82.25}$_{\pm0.46}$                             & \textit{55.84}$_{\pm1.78}$                             & \textit{87.71}$_{\pm4.25}$                     & \textit{84.18}$_{\pm0.19}$     \\
\textbf{\qquad+ UMLS20}                              & \textit{\textbf{83.26}}$^\dagger_{\pm0.32}\uparrow$  & \textit{62.84}$^\dagger_{\pm2.71}\uparrow$             & \textit{90.64}$^\dagger_{\pm2.43}\uparrow$     & \textit{\textbf{84.70}}$_{\pm0.19}\uparrow$     \\ 
\textbf{\qquad+ Onto20Type}                  & 82.17$_{\pm0.62}$                           & 55.40$_{\pm5.57}$                             & 86.36$_{\pm3.07}$                     & 83.94$_{\pm0.63}$     \\
\textbf{\qquad+ Onto20Fused}              & 82.39$_{\pm0.65}\uparrow$                   & 56.12$_{\pm2.91}\uparrow$                     & 84.36$_{\pm4.73}$                     & 83.97$_{\pm0.59}$     \\ \hdashline
\textbf{BioLinkBERT-base}                & 82.21$_{\pm0.87}$                             & 56.76$_{\pm3.00}$                             & 91.29$_{\pm3.18}$                     & 84.1$_{\pm0.03}$      \\
\textbf{\qquad+UMLS20}                      & 82.36$_{\pm0.57}\uparrow$                   & 63.62$^\dagger_{\pm5.31}\uparrow$             & 91.50$_{\pm2.25}\uparrow$             & 83.78$_{\pm0.09}$     \\
\textbf{\qquad+Onto20Type}               & 82.37$_{\pm0.42}\uparrow$                   & 60.46$_{\pm5.81}\uparrow$                     & \textbf{92.14}$_{\pm2.30}\uparrow$    & 82.84$_{\pm0.34}$     \\
\textbf{\qquad+Onto20Fused}          & 82.24$_{\pm1.25}\uparrow$                   & 63.28$^\dagger_{\pm4.46}\uparrow$             & 90.57$_{\pm3.14}$                     & 83.69$_{\pm0.55}$     \\ \hline
\end{tabular}
\caption{Final results of the model experiments: The metric for HoC is Micro F1, while for the other three it is accuracy. The best results for every task are in bold. "$\uparrow$" denotes that improvements are observed when compared to the base model. “$\dagger$" denotes a statistically significant better result over the base model (T-test, p $<$ 0.05). The results in \textit{italic} are taken from previous works, while the rest of results comes from our experiments.
%For all MoP metrics, we took the S20Rel metric for better comparability. Because of their unclarified status, we excluded all original BioLinkBERT results in favor of our reproduced results. DAKI and KEBLM did not evaluate on all of our used benchmarks and did not provide standard deviation or p-tests, but we still included their results for completeness.
} \label{tab:final_results}
\end{table*}

\subsection{Numerical Analysis}

Table \ref{tab:final_results} shows the final results of the experiments. Each section first shows the performance of the base biomedical model on its own, namely SciBERT \cite{scibert}, BioBERT \cite{biobert}, PubMedBERT \cite{blurb}, and BioLinkBERT \cite{biolinkbert}. Afterwards, indentended rows show the performance of knowledge-enhanced versions of the models. For SciBERT and BioBERT, we report on competing approaches that use structured knowledge integration: MoP \cite{meng-etal-2021-mixture}, DAKI \cite{daki}, and KEBLM \cite{keblm}. For PubMedBERT and BioLinkBERT, we report on the knowledge-enhanced versions as described in our paper, augmented with structured knowledge from knowledge graphs UMLS20, Onto20Fused, and Onto20Type. It should be noted that the BioLinkBERT results differ from the ones in the original publication because we report on averaged experiment results over multiple runs, unlike the best single run in the original paper.

The results demonstrate that our knowledge enhancement approach improved PubMedBERT in six instances and the BioLinkBERT model in eight instances, either with the UMLS data or the OntoChem data. Notably, there is a difference in the margin of improvement between the datasets. For HoC, the improvement is either negligible or 1\% in the best case. This shows that the task of trying to classify document abstracts according to cancer properties is mostly dependent on the document context itself and does not noticeably benefit from external knowledge. Similar is the case for MedNLI, which either deteriorates or improves less than 1\%, showing that entailment recognition is mostly tied to the reasoning capabilities of a language model and not the deeper medical knowledge. 

On the other hand, the two question-answering datasets experience noticeable improvements. This makes sense considering the knowledge-intensive nature of QA, where factual knowledge is at its core.  Especially for PubMedQA, both base PLMs get a 7\% jump in accuracy with different KGs. An impressive result is the BioLinkBert-base + Onto20Type model achieving state-of-the-art performance on the BioASQ7b dataset (when looking at the averaged performance over 10 runs). When looking at the difference between the two styles of OntoChem relations, the fused version was superior for PubMedQA (by 3\%), while the more detailed, typed version performed better for BioASQ (by 1.5\%). We attribute this to the slight difference in the domain of these two datasets -- BioASQ contains more questions relating to chemical knowledge, where specific types could come into play, while PubMedQA covers diverse medical diagnoses and treatments. 

An interesting result that we have to investigate further is the relatively worse performance of our approach with OntoChem KGs on PubMedBERT compared to BioLinkBERT, even when factoring in the stronger base performance of BioLinkBERT. When the base models don’t match, it is hard to distinguish whether performance gains or losses come from the difference in base models or the difference in the adapter-based approaches. Here, the base models of BioLinkBERT generally perform better than those of PubMedBERT or SciBERT over a variety of tasks. Therefore, whenever we use BioLinkBERT, we cannot say how much of the performance gains come from the superiority of our approach versus the superiority of the base model.

\subsection{Qualitative Analysis}

To investigate the performance of our knowledge-enhanced models on a deeper level, we decided to look at the classification performance on an instance level and singled out some interesting examples. Table \ref{tab:qualitative} shows two instances from the BioASQ dataset where our knowledge-enhanced model predicted the answer correctly, unlike the base model. Instances in BioASQ consist of a question and context, and the goal is to answer the question with a yes/no verdict. 

\begin{table*}[!htpb]
\centering
\begin{tabular}{p{35mm} p{60mm} p{37mm} p{6mm} }
\hline
 \textbf{Question }      &   \textbf{Context}         &  \textbf{Predictions}   &  \\ \hline
Can Diazepam be beneficial in the treatment of traumatic brain injury?                 &   The present experiment examined the effects of diazepam, a positive modulator at the GABA(A) receptor, on survival and cognitive performance in traumatically brain-injured animals.      &  \textbf{BioLinkBERT}:  \newline \textbf{BLBERT+Onto20Type}: \newline \hfill \newline \textbf{Gold Label}: &  \textbf{\textcolor{red}{no}} \newline \textbf{\textcolor{teal}{yes}} \hfill \newline \newline \textbf{\textcolor{teal}{yes}} \\
                     &                    &           &      \\ 

Does axitinib prolong the survival of pancreatic cancer patients?                  &  Axitinib/gemcitabine, while tolerated, did not provide survival benefit over gemcitabine alone in patients with advanced pancreatic cancer from Japan or other regions [...]. & \textbf{BioLinkBERT}:  \newline \textbf{BLBERT+Onto20Type}: \newline \hfill \newline \textbf{Gold Label}: &  \textbf{\textcolor{teal}{yes}} \newline \textbf{\textcolor{red}{no}} \newline  \hfill \newline \textbf{\textcolor{red}{no}} \\
                
  \hline

\end{tabular}
\caption{\label{tab:qualitative} Examples of two instances from the BioASQ dataset (with a question, context, and verdict) where the knowledge-enhanced model performed correctly, unlike its vanilla counterpart.}
\end{table*}

\iffalse
%TODO: table
\begin{table}[htpb]
\centering
\small
\begin{tabular}{p{\columnwidth}}
\\
\hline
\textbf{Question:} \small Can Diazepam be beneficial in the treatment of traumatic brain injury? \\
\hline
\textbf{Context:} \small The present experiment examined the effects of diazepam, a positive modulator at the GABA(A) receptor, on survival and cognitive performance in traumatically brain-injured animals. \\
\hline
\textbf{Prediction:}  BioLinkBERT+Onto20Type: yes || BioLinkBERT: no || Gold Label: yes \\
\hline
\\
\hline
\textbf{Question:} \small Does axitinib prolong the survival of pancreatic cancer patients? \\
\hline
\textbf{Context:} \small Axitinib/gemcitabine, while tolerated, did not provide survival benefit over gemcitabine alone in patients with advanced pancreatic cancer from Japan or other regions [...]. \\
\hline
\textbf{Prediction:}  BioLinkBERT+Onto20Type: no || BioLinkBERT: yes || Gold Label: no \\
\hline
\end{tabular}

\caption{Examples of two instances from the BioASQ dataset (with a question, context, and verdict) where the knowledge-enhanced model performed correctly, unlike its vanilla counterpart.}
\label{tab:qualitative}

\end{table}

\fi

The first row contains a question on the relationship between Diazepam and traumatic brain injury. While the vanilla BioLinkBERT answered the question incorrectly, our knowledge-enhanced BioLinkBERT + Onto20Type model gave the correct answer. Diazepam (first marketed as Valium) is listed as an entity in the OntoChem KG, where it has a direct relation to brain injuries -- the full triple is "diazepam [substance] treats [disease] brain injury" (see also figure \ref{fig:ontochem}. It is likely that, thanks to the injection of this knowledge, the enhanced model was able to deduce the answer, while the base model was not.   

The second row shows a question about axitinib and its relation to pancreatic cancer. Here, the base version of BioLinkBERT incorrectly predicted that axitinib does prolong the survival of pancreatic cancer patients, while our BioLinkBERT + Onto20Type model gave the correct negative answer. This time, there is no relation between axitinib and any form of cancer listed in the KG. Therefore, our enhanced model might have been able to rely on its injected knowledge and deduce that there are no such connections between the entities in question.

\section{\uppercase{Conclusion}}

This paper investigated the performance of biomedical pre-trained language models when enhanced with structured domain-specific biomedical knowledge. For this purpose, we utilized two biomedical PLMs (PubMedBERT and BioLinkBERT) and external knowledge from two large KGs,  UMLS and OntoChem. The KGs were partitioned into smaller subgraphs and later fused into a common knowledge representation. The knowledge was injected into the PLMs by using lightweight but powerful adapter modules. We tested the performance on four downstream biomedical NLP tasks and showed that the knowledge-enhanced models consistently improved the results, indicating a clear benefit of infusing external structured knowledge into unstructured PLMs. By updating the adapter weights, which are only about 1--2\% amount of PLM weights, the performance (in best setting) increased on HoC and MedNLI for 1\%, on BioASQ 3\%, and on PubMedQA 7\%. This demonstrates the power of using adapter modules to fine-tune PLMs for domain-specific purposes. Moreover, we have demonstrated that OntoChem is a viable alternative to UMLS and other knowledge sources in the field of biomedical knowledge enhancement.

\paragraph{Future Research}
In future work, we would like to further investigate the potential of the OntoChem ontology. Besides entities and relations, every data triple comes with the source sentence from which it was extracted. Drawing inspiration from works like K-Adapter \cite{kadapter}, this linguistic knowledge could be extracted and used in additional adapters to enhance the models. Moreover, the idea of merging together the data from Ontochem with subgraphs from MSI \cite{msi}, UMLS \cite{bodenreider2004unified}, or PubChem\cite{pubchem} presents a promising direction. Finally, future work could be more human-centric and have medical professionals curate the KGs. This way, the resulting KELMs would be tailored directly by those who use them. 

\paragraph{Limitations}
Our research did not come without certain challenges and limitations. A portion of the data from OntoChem was not usable due to incomplete ID mappings. As a result, only a fraction of the available knowledge was integrated into the experimental segment of this work, which has likely led to less thoroughly connected KGs. Additionally, medical professionals often indicate concerns regarding ethical questions and the development and use of LLMs in bio-medicine. While our methodology and models will likely not be used in practice without further research and improvements, we did not specifically address the medical community's concerns in our work. We tried to improve the overall model performance and factual accuracy to reduce hallucinations, but there is no way to entirely eliminate the risk of wrong predictions and other critical issues. At the time of writing, we are conducting a survey involving clinicians to address their concerns in our future work. 

\bibliographystyle{apalike}
{\small
\bibliography{example}}

\end{document}